\documentclass[conference]{IEEEtran}

\usepackage{cite}
\usepackage{amsmath,amssymb,amsfonts}
\usepackage{algorithmic}
\usepackage{graphicx}
\usepackage{textcomp}
\usepackage{xcolor}
\usepackage{booktabs}
\usepackage{graphicx}
\usepackage{textcomp}
\usepackage{url}
\usepackage[table]{xcolor} 
\def\BibTeX{{\rm B\kern-.05em{\sc i\kern-.025em b}\kern-.08em
    T\kern-.1667em\lower.7ex\hbox{E}\kern-.125emX}}
\begin{document}

\title{CTForensics: A Comprehensive Dataset and Method for AI-Generated CT Image Detection}


\author{
\IEEEauthorblockN{Yiheng Li$^{1,2}$, Zichang Tan$^{4}$, Guoqing Xu$^{1,2}$, Yichun Yeh$^{1,2}$, Yang Yang$^{1,2,\ast}$, Zhen Lei$^{1,2,3}$}
\IEEEauthorblockA{
$^1$ School of Artificial Intelligence, University of Chinese Academy of Sciences, Beijing, China\\
$^2$ MAIS, Institute of Automation, Chinese Academy of Sciences, Beijing, China\\
$^3$ CAIR, HKISI, Chinese Academy of Sciences, Beijing, China\\
$^4$ Sangfor Technologies Inc., Shenzhen, China \quad $\ast$ Corresponding author\\
\textit{Emails:} \texttt{\{liyiheng2024, yangyang2013, zhen.lei\}@ia.ac.cn, tanzichang@foxmail.com}
}}
\maketitle

\author{\IEEEauthorblockN{1\textsuperscript{st} Given Name Surname}
\IEEEauthorblockA{\textit{dept. name of organization (of Aff.)} \\
\textit{name of organization (of Aff.)}\\
City, Country \\
email address or ORCID}
\and
\IEEEauthorblockN{2\textsuperscript{nd} Given Name Surname}
\IEEEauthorblockA{\textit{dept. name of organization (of Aff.)} \\
\textit{name of organization (of Aff.)}\\
City, Country \\
email address or ORCID}
}

\begin{abstract}
Recent advances in generative AI have made synthetic Computed Tomography (CT) images increasingly realistic, enabling promising applications in medical data augmentation while raising serious concerns about clinical safety and data trustworthiness. Detecting AI-generated CT images remains challenging for two key reasons: existing benchmarks cover only limited generation sources, and many detectors are adapted from natural-image forensics without explicitly modeling CT-specific imaging properties. In this paper, we introduce \textbf{CTForensics}, a dataset for detecting AI-generated CT images. CTForensics contains 75,990 2D CT images, including a dedicated test benchmark of 29,990 balanced authentic and generated samples from ten representative CT generative models spanning GAN-based and diffusion-based paradigms. We further propose the \textbf{E}nhanced \textbf{S}patial-\textbf{F}requency \textbf{CT} \textbf{F}orgery \textbf{D}etector \textbf{(ESF-CTFD)}, a CT-oriented CNN framework built around a Wavelet-Enhanced Central Stem, Multi-Scale Spatial Aggregation, and a Frequency-Aware Prediction Block. The Wavelet-Enhanced Central Stem enhances local intensity correlations and high-frequency residuals, Multi-Scale Spatial Aggregation aligns anatomical features across resolutions with lightweight residual units, and the Frequency-Aware Prediction Block models global spectral artifacts. Extensive experiments on CTForensics show that ESF-CTFD achieves 96.01\% mAcc and 99.96\% mAP, outperforming existing methods and maintaining strong robustness under realistic perturbations with only a 0.99\% average drop.
Codes will be available at \url{https://github.com/liyih/CTForensics}.
\end{abstract}


\section{Introduction}
In recent years, rapid advances in generative artificial intelligence (AI), particularly generative adversarial networks (GANs) \cite{goodfellow2014generative} and diffusion models \cite{song2020score}, have facilitated the synthesis of highly realistic images. In medical imaging, these models have been increasingly leveraged to generate synthetic data for applications such as data augmentation and cross-modality translation, with computed tomography (CT) serving as a representative example \cite{mirsky2019ct, jiang2025lung}. However, the increasing visual fidelity of AI-generated CT images raises serious concerns regarding clinical safety and data trustworthiness. Synthetic scans may be challenging for both human experts and automated systems to distinguish from authentic scans, potentially leading to erroneous clinical interpretations or diagnoses. Therefore, developing robust detection mechanisms that can reliably discriminate between authentic and AI-generated CT images is essential for safeguarding the reliability and integrity of medical diagnostics.

\begin{figure}[t]
  \centering
   \includegraphics[width=0.48\textwidth]{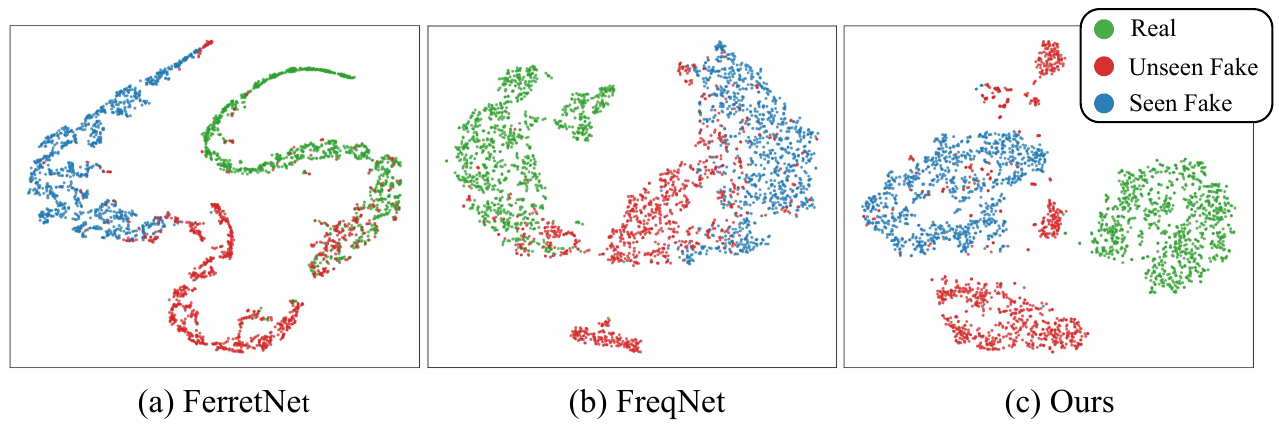}
   \caption{\textbf{t-SNE visualization of feature embeddings.} Red, blue, and green denote unseen fake, seen fake, and real CT slices, respectively. Compared with FreqNet and FerretNet, ESF-CTFD forms clearer clusters, indicating stronger discrimination between authentic and generated CT images.}
   \label{fig:tsne}
\end{figure}

Despite the urgent need for effective countermeasures, a primary obstacle in the field is the lack of a CT forgery benchmark that reflects realistic and diverse generation scenarios. Existing datasets \cite{albahli2024mednet,li2025toward} only contain fake CT scans from limited kinds of CT generative methods. To bridge this gap, we propose \textbf{CTForensics}, which separates a training split from a dedicated test benchmark. The test benchmark contains 29,990 2D CT images with balanced authentic and generated samples, covering ten representative CT generative models, including four GAN-based methods (CGAN \cite{loey2025deep}, CTGAN \cite{mirsky2019ct}, sRD-GAN \cite{lee2022diverse}, and HA-GAN \cite{sun2022hierarchical}) and six diffusion-based methods (GenerateCT \cite{hamamci2024generatect}, LungDDPM \cite{jiang2025lung}, LungDDPM+ \cite{jiang2025lung+}, MedSyn \cite{xu2024medsyn}, StableDiffusion \cite{rombach2022high}, and Text2CT \cite{molino2025text}). The generated images are collected from public releases or produced with official checkpoints, while their authentic counterparts are sampled from CT-RATE \cite{hamamci2024developing} with matched quantities for each test source. All images are standardized as single-channel grayscale PNGs and checked through automated and manual quality review. This design enables systematic evaluation across heterogeneous generators and provides a challenging benchmark for CT forgery detection. 

\begin{figure*}[t]
  \centering
   \includegraphics[width=\textwidth]{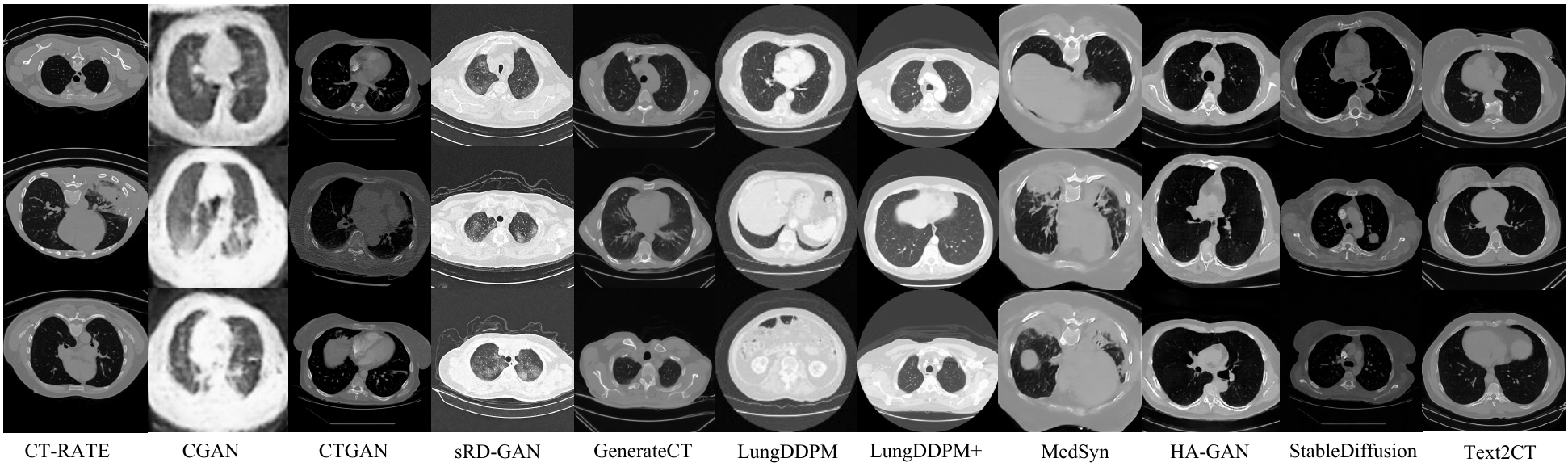}
   \caption{\textbf{Visualization of CTForensics.} Examples of authentic CT slices and AI-generated slices from representative GAN-based and diffusion-based generators.}
   \label{fig:data_vis}
\end{figure*}

Another significant challenge lies in the methodology. Existing forgery detection methods \cite{ojha2023towards,tan2024rethinking,li2025improving,li2026towards,zheng2024breaking,tan2024frequency,liang2025ferretnet} are largely adapted from natural image analysis and often overlook the imaging characteristics of medical CT scans. Unlike RGB images, CT slices are single-channel tomographic images whose forensic cues are often reflected in subtle grayscale attenuation patterns, local intensity correlations, anatomical boundaries, and spectral statistics. Although wavelet and frequency modeling have also been explored in natural-image forensics \cite{tan2024frequency}, directly transplanting them is insufficient because CT detection requires enhancing weak residual artifacts while preserving anatomical structure. To this end, we propose \textbf{ESF-CTFD}, a CT-oriented framework organized around a Wavelet-Enhanced Central Stem, Multi-Scale Spatial Aggregation, and prediction by a Frequency-Aware Prediction Block. Specifically, the Wavelet-Enhanced Central Stem combines wavelet decomposition with central correlation convolution to emphasize directional high-frequency residuals and local intensity relationships in CT slices. The Multi-Scale Spatial Aggregation module progressively aligns and aggregates features across adjacent scales to retain anatomical consistency, with Spatial Residual Blocks serving as lightweight units for this process. The Frequency-Aware Prediction Block further models global spectral artifacts after CT-aware spatial-wavelet enhancement. This design aligns the detector with CT imaging properties rather than relying on generic RGB-domain texture cues. 

Extensive experiments on CTForensics demonstrate that ESF-CTFD consistently outperforms state-of-the-art methods and maintains strong robustness under realistic perturbations. As shown in Fig. \ref{fig:tsne}, ESF-CTFD produces more clearly separated t-SNE embeddings \cite{van2008visualizing} than FreqNet and FerretNet, especially for distinguishing real CT slices from unseen generated samples. The main contributions are as follows:
\begin{itemize}
    \item We propose \textbf{CTForensics}, a CT forgery detection resource whose test benchmark contains 29,990 images from ten representative generative methods.
    \item We design \textbf{ESF-CTFD}, a CT-oriented detector built on a Wavelet-Enhanced Central Stem, Multi-Scale Spatial Aggregation, and a Frequency-Aware Prediction Block to jointly model wavelet-enhanced local cues, cross-scale anatomical structure, and frequency-domain artifacts.
    \item We provide comprehensive systematic evaluations showing remarkable generalization capability and strong robustness for the task of AI-generated CT image detection.
\end{itemize}

\section{Related Work}
\subsection{CT Image Generation.}
Recent advances in generative models have enabled the synthesis of realistic CT images for data augmentation and simulation.
Some work \cite{lee2022diverse,loey2025deep,touati2021feature} uses GAN-based \cite{goodfellow2014generative} networks to synthesize CT images. For instance, CTGAN \cite{mirsky2019ct} injects or removes cancerous tumors using a conditional GAN. 
HA-GAN \cite{sun2022hierarchical} proposed an end-to-end architecture that can simultaneously generate a low-resolution version of an image and a randomly selected sub-volume from the high-resolution image.
In contrast, some methods \cite{jiang2025lung+,lyu2022conversion,shao2025trace,zhu2024generative} adopt diffusion-based \cite{song2020score} architectures to generate CT images. Specifically, Lung-DDPM \cite{jiang2025lung} generates high-fidelity CT images based on semantic layout-guided denoising diffusion probabilistic models. 
BTD \cite{grabovski2025back} leverages StableDiffusion \cite{rombach2022high} for medical image in-painting. It fine-tunes the latent diffusion model on medical scans to inject or remove tumor evidence, producing localized tampering cases with subtle anatomical changes.
Recent methods \cite{xu2024medsyn} generate CT images based on medical language text prompts. For example, GenerateCT \cite{hamamci2024generatect} generates CT images conditioned on medical text prompts using a causal vision transformer and a text-conditional super-resolution diffusion model.
Text2CT \cite{molino2025text} generates anatomically consistent 3D CT volumes from radiology text prompts using a latent diffusion model with 3D contrastive vision-language pretraining and a volumetric VAE.

\begin{table*}[t]
\caption{\textbf{Statistics of the CTForensics training/testing splits.} Real \# and Fake \# denote the numbers of authentic CT slices sampled from CT-RATE and generated CT slices from each source, respectively. Protocol indicates whether a source is used for training, seen-generator testing, or unseen-generator testing.}
\centering
\scriptsize
\resizebox{\textwidth}{!}{
\begin{tabular}{c c c c c c c c}
\toprule
Split & Generator & Type & Real \# & Fake \# & Generation setting & Acquisition & Protocol \\
\midrule
Training & HA-GAN \cite{sun2022hierarchical} & GAN & 23,000 & 23,000 & Hierarchical synthesis & Official checkpoint & Training \\
Testing & HA-GAN \cite{sun2022hierarchical} & GAN & 2,000 & 2,000 & Hierarchical synthesis & Official checkpoint & Seen test \\
Testing & CGAN \cite{loey2025deep} & GAN & 2,000 & 2,000 & Class-conditional synthesis & Public release & Unseen test \\
Testing & CTGAN \cite{mirsky2019ct} & GAN & 1,622 & 1,622 & Tumor in-painting & BTD test set & Unseen test \\
Testing & sRD-GAN \cite{lee2022diverse} & GAN & 367 & 367 & Image translation & Public release & Unseen test \\
Testing & GenerateCT \cite{hamamci2024generatect} & Diffusion & 2,000 & 2,000 & Text-to-CT synthesis & Released 3D volumes & Unseen test \\
Testing & LungDDPM \cite{jiang2025lung} & Diffusion & 384 & 384 & Semantic-layout guidance & Official checkpoint & Unseen test \\
Testing & LungDDPM+ \cite{jiang2025lung+} & Diffusion & 1,000 & 1,000 & Semantic-layout guidance & Official checkpoint & Unseen test \\
Testing & MedSyn \cite{xu2024medsyn} & Diffusion & 2,000 & 2,000 & Text-to-CT synthesis & Official checkpoint & Unseen test \\
Testing & StableDiffusion \cite{rombach2022high} & Diffusion & 1,622 & 1,622 & Tumor in-painting & BTD test set & Unseen test \\
Testing & Text2CT \cite{molino2025text} & Diffusion & 2,000 & 2,000 & Text-to-CT synthesis & Released 3D volumes & Unseen test \\
\bottomrule
\end{tabular}
}
\label{tab:benchmark_stats}
\end{table*}

\subsection{Forgery CT Detection.}
To ensure the privacy and security of users, some work \cite{prezja2022deepfake,solaiyappan2022machine} has begun to focus on CT forgery detection. MedForensics \cite{li2025toward} proposes a large-scale dataset that discusses six medical modalities, while BTD \cite{grabovski2025back} introduces a diffusion-based anomaly detection method for medical images. However, both methods share a common limitation in that they do not systematically evaluate model generalization across diverse CT generative methods. Specifically, the former considers only one generation approach \cite{konz2024anatomically}, while the latter examines only two \cite{mirsky2019ct,rombach2022high}. Most methods conduct a standard convolutional network \cite{he2016deep} or their variants to extract discriminative features for CT forgery detection. For example, MedNet \cite{albahli2024mednet} proposes a customized EfficientNetV2-based \cite{tan2021efficientnetv2} model with a spatial–channel attention mechanism to detect lung CT deepfakes. 
However, these architectures are insufficient to fully capture forgery artifacts in CT images. In contrast, to fully address the CT forgery detection task, we first propose a comprehensive dataset named CTForensics, which contains ten diverse generative methods. Moreover, we introduce ESF-CTFD, which combines a Wavelet-Enhanced Central Stem, Multi-Scale Spatial Aggregation, and a Frequency-Aware Prediction Block to capture CT-specific forgery artifacts.

\section{CTForensics Dataset}
\subsection{Task Definition}
Given a CT slice $I \in \mathbb{R}^{C \times H \times W}$, where $C$, $H$, and $W$ denote the number of channels, height, and width, respectively, the goal of AI-generated CT detection is to learn a binary classifier $f_{\theta}$ that distinguishes authentic slices from AI-generated ones. The label is denoted as $y \in \{0,1\}$, with $y=0$ for authentic slices and $y=1$ for AI-generated slices.
 We use 2D slices rather than full 3D volumes for two reasons. First, existing CT generators release data in different formats, including slices, volumes, and local in-painting results, so slice-level standardization enables fair comparison across heterogeneous sources. Second, clinical and forensic workflows often inspect CT studies as slice sequences, making slice-level detection a practical unit when complete volumes are unavailable.

Beyond standard binary classification, CTForensics emphasizes out-of-domain evaluation. The detector is trained with labeled authentic and generated CT slices from the training split, and is then tested on multiple subsets produced by heterogeneous CT generative models. Most test generators are unseen during training and differ in architecture, training data, generation target, and conditioning strategy. This protocol directly measures whether a detector can generalize to unseen generators and capture transferable CT forgery cues, instead of overfitting to artifacts tied to a specific source.

\subsection{Dataset Construction}
To support this evaluation, we construct \textbf{CTForensics} with a clear separation between the training split and the test benchmark. The training split is used only to fit detectors and contains 23,000 authentic CT slices from CT-RATE \cite{hamamci2024developing} and 23,000 generated slices from HA-GAN \cite{sun2022hierarchical}. The core benchmark refers to the testing split, which contains 14,995 authentic slices and 14,995 generated slices from ten CT generative models. For each testing generator, the number of authentic CT slices is matched to the corresponding generated samples, enabling balanced source-wise evaluation. The full train/test statistics are summarized in Tab. \ref{tab:benchmark_stats}.

\begin{figure*}[t]
  \centering
   \includegraphics[width=\textwidth]{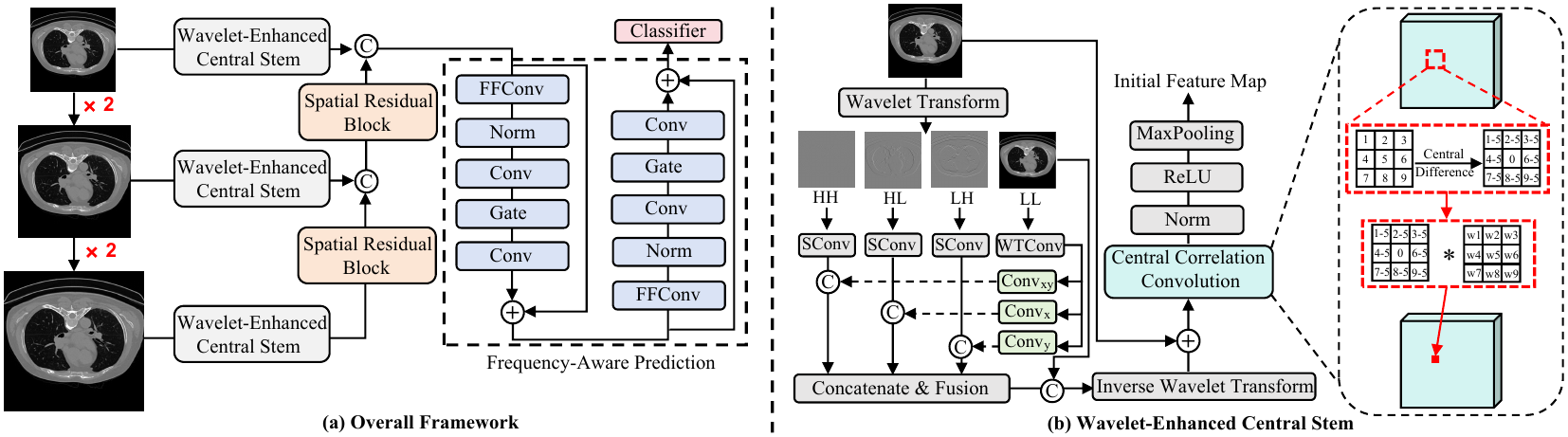}
   \caption{\textbf{Overall framework of our proposed ESF-CTFD.} The input CT slice is processed at three resolutions. Each branch first applies a Wavelet-Enhanced Central Stem to strengthen local intensity correlations and high-frequency residuals. Multi-Scale Spatial Aggregation progressively aligns and concatenates features across resolutions with lightweight Spatial Residual Blocks, then a Frequency-Aware Prediction Block refines the merged representation for final authenticity prediction.}
   \label{fig2}
\end{figure*}

Rather than focusing on a single synthesis pattern, the testing split is designed to cover diverse generation scenarios, including class-conditional generation, tumor in-painting, image translation, hierarchical synthesis, semantic-layout guidance, and Text-to-CT synthesis. HA-GAN appears in both training and testing as a seen-generator reference, while the other generators appear only in testing. Therefore, the main evaluation focuses on out-of-domain performance on unseen generators rather than closed-set recognition.

The generated samples are collected from both public releases and official model inference. CGAN, CTGAN, StableDiffusion, sRD-GAN, GenerateCT, and Text2CT are obtained from publicly released synthetic CT data; among them, CTGAN and StableDiffusion are from the BTD test set \cite{grabovski2025back}, and GenerateCT and Text2CT are converted from released 3D volumes by extracting 2D slices. HA-GAN, LungDDPM, LungDDPM+, and MedSyn are generated with official pretrained checkpoints. For MedSyn, prompts are sampled from CT-RATE \cite{hamamci2024developing} radiology reports. Finally, all samples are converted into a unified slice-level format, saved as single-channel grayscale PNG files, and checked through automated screening followed by manual review. Examples of authentic and generated CT slices are shown in Fig. \ref{fig:data_vis}.

\section{Method}
\subsection{Overview}
As shown in Fig. \ref{fig2}, ESF-CTFD is a CT-oriented detector that combines wavelet-guided local enhancement, Multi-Scale Spatial Aggregation, and prediction by a Frequency-Aware Prediction Block.
For a CT slice $I$, we construct three multi-scale inputs $\mathcal{I}=\{I^{(4)}, I^{(2)}, I^{(1)}\}$, where $I^{(4)}$, $I^{(2)}$, and $I^{(1)}$ correspond to scale factors $4$, $2$, and $1$, respectively.
Each resolution is processed by a Wavelet-Enhanced Central Stem, which enhances CT-specific residual cues before conventional feature extraction. The resulting features are progressively aligned and concatenated across scales, where Spatial Residual Blocks are used as lightweight residual units inside the aggregation path. A Frequency-Aware Prediction Block based on Fast Fourier Convolution models global spectral artifacts, and a linear classification head predicts the authenticity score.

This design follows two considerations. First, generated CT slices often preserve plausible anatomy but leave weak local residuals around intensity transitions, lesion-like regions, or reconstruction boundaries. Therefore, the early stage should emphasize local grayscale correlations and directional high-frequency responses. Second, different generators introduce artifacts at different spatial extents, ranging from subtle texture irregularities to global spectrum shifts. ESF-CTFD therefore aggregates wavelet-enhanced features across multiple resolutions and further refines them in the frequency domain.

\subsection{Wavelet-Enhanced Central Stem}
For each input scale, we apply a wavelet enhancement module before the central stem. Given a scaled input $x \in \mathcal{I}$, the discrete wavelet transform (DWT) decomposes $x$ into one low-frequency sub-band and three high-frequency sub-bands:
\begin{equation}
\label{eq1}
LL, HL, LH, HH = \operatorname{DWT}(x).
\end{equation}
Here, $LL$ denotes the low-low sub-band, which primarily preserves anatomical structures and smooth attenuation patterns, whereas $HL$, $LH$, and $HH$ denote the high-low, low-high, and high-high sub-bands that capture directional residual details. To preserve structural context, the $LL$ sub-band is enhanced using stacked wavelet convolution layers (WTConv) \cite{finder2024wavelet,zhang2025cwnet}, which progressively expand the receptive field and produce $LL^{\prime}$. In parallel, the three high-frequency sub-bands are processed by independent depth-wise separable convolution layers (SConv) \cite{chollet2017xception}, suppressing redundant channel interactions while retaining local residual responses to obtain $HL^{\prime}$, $LH^{\prime}$, and $HH^{\prime}$, respectively.

To make the high-frequency enhancement aware of CT structures, we further derive directional responses from the enhanced low-frequency component. Horizontal, vertical, and diagonal filters are applied to $LL^{\prime}$ and injected into the corresponding high-frequency branches:

\begin{equation}
\label{eq2}
\left\lbrace
\begin{aligned}
HL^{e} &= [HL^{\prime}; \operatorname{Conv}_{x}(LL^{\prime})], \\
LH^{e} &= [LH^{\prime}; \operatorname{Conv}_{y}(LL^{\prime})], \\
HH^{e} &= [HH^{\prime}; \operatorname{Conv}_{xy}(LL^{\prime})].
\end{aligned}
\right.
\end{equation}
Here, $\operatorname{Conv}_{x}$, $\operatorname{Conv}_{y}$, and $\operatorname{Conv}_{xy}$ are fixed $3\times3$ horizontal, vertical, and diagonal kernels, respectively. $[.;.]$ denotes channel concatenation. 
These enhanced sub-bands are reconstructed and added to the original input to obtain the enhanced input $x^{e}$ via the following process:
\begin{equation}
\label{eq3}
x^{e}=\operatorname{IDWT}([LL^{\prime}; \Phi_f([HL^{e}; LH^{e}; HH^{e}])]) + x,
\end{equation}
where $\Phi_f$ is a fusion function composed by convolution, normalization, and a feed-forward layer, and $\operatorname{IDWT}(\cdot)$ denotes the inverse discrete wavelet transform,

After wavelet enhancement, a central correlation convolution \cite{yu2020searching} is used as the stem operator to further emphasize local residual patterns in CT slices. Unlike standard convolution, it subtracts the center pixel from its neighbors before aggregation, making the response more sensitive to subtle grayscale discontinuities and manipulation traces:
\begin{equation}
\label{eq4}
\begin{aligned}
y(p_0) &= \sum_{p_i \in R^l} w(p_i) \cdot (x^e(p_{0+i}) - x^e(p_{0})) \\
&= \sum_{p_i \in R^l} w(p_i) \cdot x^e(p_{0+i}) - x^e(p_0) \cdot \sum_{p_i \in R^l} w(p_i),
\end{aligned}
\end{equation}
where $p_0$ is the center pixel, $p_i$ indexes neighboring pixels in the local region $R^l$, $x^e(\cdot)$ denotes the pixel value, and $w(p_i)$ is the convolution weight at location $p_i$. This operation is suitable for CT forensics because it explicitly models local grayscale differences rather than RGB texture patterns. The stem output is then normalized, activated, and downsampled by max pooling.

\begin{figure*}[t]
  \centering
   \includegraphics[width=\textwidth]{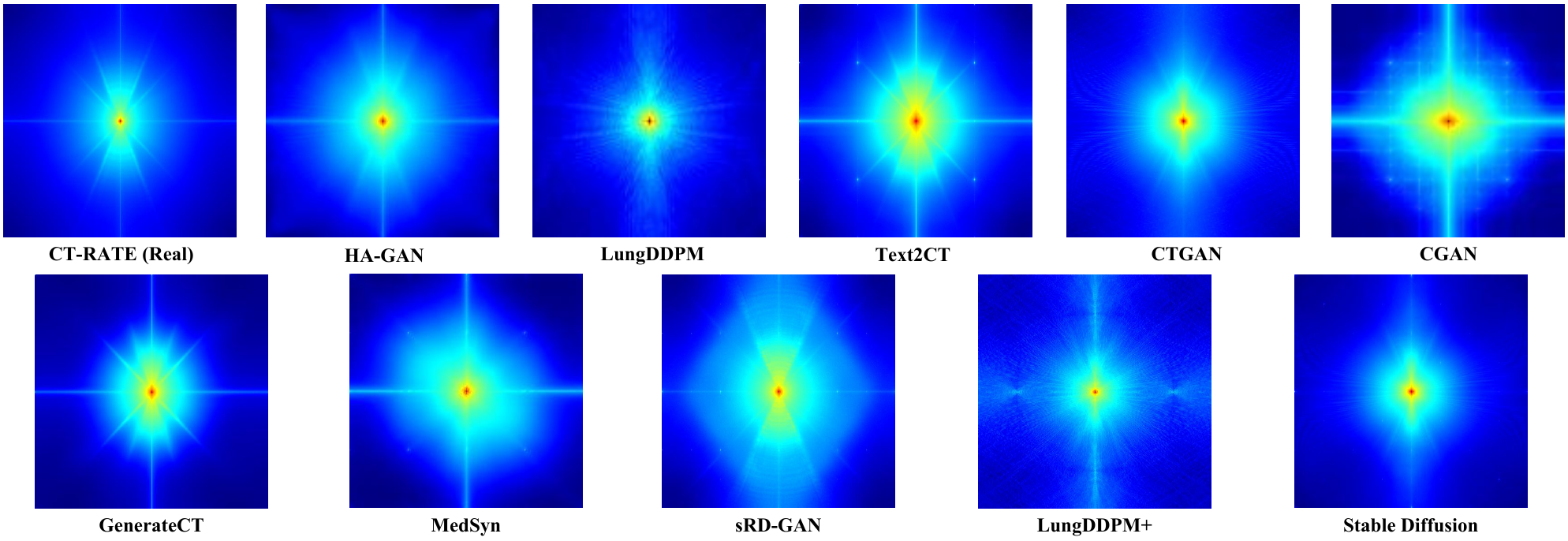}
   \caption{\textbf{Frequency analysis of authentic and AI-generated CT slices.} Averaged spectra are visualized to show the distributional differences between real CT images and generated samples in the frequency domain, motivating the Frequency-Aware Prediction Block.}
   \label{fig3}
\end{figure*}

\subsection{Multi-Scale Spatial Aggregation}
ESF-CTFD uses three resolution branches to capture artifacts at different spatial extents. In the implementation, the input slice is resized to $4\times$, $2\times$, and $1\times$ resolutions. Each branch has an independent Wavelet-Enhanced Central Stem, producing wavelet-enhanced stem features with the same initial channel width. Starting from the highest-resolution branch, the multi-scale aggregation path progressively downsamples the feature map and aligns it with the next lower-resolution branch through Spatial Residual Blocks. The aligned feature and the stem feature from the lower-resolution branch are concatenated along the channel dimension. This process is repeated until the three branches are merged.

Each spatial residual block is built upon ResNet-style residual units \cite{he2016deep}. It consists of stacked convolution, batch normalization, and ReLU operations. When spatial downsampling is required, a stride-2 residual path is adopted to match feature resolutions. The residual connection helps preserve stable anatomical responses during feature propagation. The stride operation further aligns neighboring resolutions for cross-scale fusion. Therefore, the spatial residual block does not introduce a complex new operator. Instead, it serves as a lightweight spatial aggregation module that bridges the outputs of the Wavelet-Enhanced Central Stem across different scales.

This multi-scale aggregation is important for CT images. High-resolution features preserve small local inconsistencies around edges and fine anatomical textures, while lower-resolution features provide more stable structural context. By aggregating them progressively across scales, ESF-CTFD avoids relying only on either local noise-like artifacts or coarse anatomical layout.

\subsection{Frequency-Aware Prediction Block}
As shown in Fig. \ref{fig3}, authentic CT slices exhibit a relatively compact and smoothly decaying spectrum concentrated around the low-frequency center. In contrast, AI-generated CT slices show generator-dependent spectral distortions, including stronger axis-aligned streaks, radial artifacts, and uneven high-frequency energy distributions. In this view, after Multi-Scale Spatial Aggregation, ESF-CTFD uses a Frequency-Aware Prediction Block to refine the merged feature before classification. As illustrated in Fig. \ref{fig2}, this block contains two residual gated Fourier paths. Each path applies Fast Fourier Convolution, followed by normalization, convolution, gate, and projection, and its output is added back to the input.


The key operator is Fast Fourier Convolution (FFConv) \cite{Chi2020FastFC}. It divides the input feature into local and global channel groups, with 75\% of channels assigned to the global branch in our implementation. The local branch uses standard spatial convolution to capture neighborhood-level responses. The global branch transforms the feature into the frequency domain, applies a $1\times1$ convolution to the real and imaginary components, and maps it back through the inverse Fourier transform. This design captures global spectral artifacts and long-range frequency correlations while preserving local spatial cues. The gate operation further improves feature selection by splitting an intermediate feature into two channel groups and multiplying them element-wise:

\begin{equation}
\label{eq5}
\operatorname{Gate}(Z)=Z_1\odot Z_2, \quad [Z_1,Z_2]=\operatorname{Split}(Z),
\end{equation}
where $Z$ is the intermediate feature and $\odot$ denotes element-wise multiplication. The first residual path uses a $5\times5$ convolution after Fast Fourier Convolution to enlarge local context, while the second path uses $1\times1$ convolutions for channel expansion and compression before prediction. Finally, the refined feature is aggregated by global average pooling and passed to a linear classifier. The model is optimized with binary cross-entropy loss.

\begin{table*}[t]
\caption{\textbf{Accuracy comparison on the CTForensics test benchmark ($\%$).} Results are reported for each generator and averaged over all ten test sources. The best-performing result for each column is \textbf{bolded}.}
 \centering
 \resizebox{\textwidth}{!}{
    \begin{tabular}{c c c c c c c c c c c| c}
    \bottomrule \hline
Methods & CGAN & CTGAN & sRD‑GAN & GenerateCT & LungDDPM & LungDDPM+ & MedSyn & HA-GAN & StableDiffusion & Text2CT & mAcc\\
\bottomrule \hline
ResNet-50 \cite{he2016deep} & 51.50 & 50.00 & 63.76 & 50.00 & 50.00 & 57.80 & \textbf{100.00} & 99.98 & 50.00 & 50.05 & 62.31\\
SAFE \cite{li2025improving} &67.00 & 54.13 & 62.81 & 50.00 & 50.13 & 50.30 & 97.13 & 99.98 & 59.49 & 50.15 & 64.11 \\
UFD \cite{ojha2023towards} & 99.88 & 50.00 & 51.09 & 50.00 & 83.72 & 70.35 & 99.15 & 99.75 & 49.91 & 50.32 & 70.42\\
NPR \cite{tan2024rethinking} & 51.88 & 73.09 & 53.13 & 50.00 & \textbf{100.00} & 82.85 & 92.83 & 99.95 & 78.39 & 50.08 & 73.22\\
FerretNet \cite{liang2025ferretnet} & 67.15 & 76.94 & 50.55 & 51.60 & \textbf{100.00} & 98.65 & 96.63 & \textbf{100.00} & 96.98 & 50.20 & 78.87\\
FreqNet \cite{tan2024frequency} &51.95 & \textbf{98.46} & 99.86 & 91.12 & 79.17 & \textbf{100.00} & 53.55 & 99.95 & \textbf{97.93} & 93.08 & 86.51\\
\bottomrule \hline
\rowcolor{blue!20} Ours & \textbf{99.20} & 89.83 & \textbf{100.00} & \textbf{99.95} & \textbf{100.00} & \textbf{100.00} & 80.75 & \textbf{100.00} & 91.12 & \textbf{99.22} & \textbf{96.01} \\
\bottomrule
\end{tabular}
}
\label{tab:1}
\end{table*}

\begin{table*}[t]
\caption{\textbf{Average precision comparison on the CTForensics test benchmark ($\%$).} Results are reported for each generator and averaged over all ten test sources. The best-performing result for each column is \textbf{bolded}.}
 \centering
 \resizebox{\textwidth}{!}{
    \begin{tabular}{c c c c c c c c c c c| c}
    \bottomrule \hline
Methods & CGAN & CTGAN & sRD‑GAN & GenerateCT & LungDDPM & LungDDPM+ & MedSyn & HA-GAN & StableDiffusion & Text2CT & mAP\\
\bottomrule \hline
ResNet-50 \cite{he2016deep} & 97.28 & 33.67 & 96.84 & 39.04 & 87.62 & 78.96 & \textbf{100.00} & \textbf{100.00} & 33.51 & 47.24 & 71.42\\
SAFE \cite{li2025improving} & 99.95 & 91.28 & 89.63 & 34.60 & 99.23 & 87.51 & 99.99 & \textbf{100.00} & 94.30 & 68.50 & 86.50\\
UFD \cite{ojha2023towards} & \textbf{100.00} & 50.12 & 84.81 & 73.56 & 99.76 & 98.60 & 99.98 & \textbf{100.00} & 66.79 & 74.58 & 84.82
\\
NPR \cite{tan2024rethinking} & 98.98 & 98.47 & 91.01 & 93.37 & \textbf{100.00} & 99.96 & 99.99 & \textbf{100.00} & 99.09 & 83.58 & 96.45\\
FerretNet \cite{liang2025ferretnet} & 99.72 & 99.05 & 80.52 & 98.96 & \textbf{100.00} & 99.99 & \textbf{100.00} & \textbf{100.00} & 99.76 & 80.95 & 95.90 \\
FreqNet \cite{tan2024frequency} &97.34 & \textbf{99.97} & \textbf{100.00} & 98.20 & 99.51 & \textbf{100.00} & 94.27 & \textbf{100.00} & \textbf{99.97} & 99.65 & 98.89\\
\bottomrule \hline
\rowcolor{blue!20}Ours & \textbf{100.00} & 99.81 & \textbf{100.00} & \textbf{100.00} & \textbf{100.00} & \textbf{100.00} & 99.91 & \textbf{100.00} & 99.89 & \textbf{99.99} & \textbf{99.96}
\\
\bottomrule
\end{tabular}
}
\label{tab:2}
\end{table*}

\section{Experiments}
\subsection{Implementation Details}
ESF-CTFD is trained on the CTForensics training split, where HA-GAN is used as the generated source, and is evaluated on the dedicated test benchmark covering ten generators. We train the model for 20 epochs with a batch size of 32. The optimizer is Adam with $\beta_1=0.9$, $\beta_2=0.999$, and an initial learning rate of $2\times10^{-4}$. A cosine annealing scheduler is applied over all training steps with a minimum learning rate of $1\times10^{-6}$. All experiments are conducted on a single NVIDIA RTX 3090 GPU.

For preprocessing, each CT slice is converted to grayscale, resized to $128\times128$, and randomly cropped to $112\times112$ during training. The network then constructs three internal resolutions, $112\times112$, $224\times224$, and $448\times448$, for Multi-Scale Spatial Aggregation. We use random horizontal flipping, Gaussian blur with probability 0.5, and JPEG compression with probability 0.5 as training augmentations. At test time, images are resized and center-cropped using the same base size. We report accuracy and average precision for each generator and their unweighted mean across the ten test generators; accuracy is computed with a threshold of 0.5. Compared detectors are evaluated under the same test protocol.

\begin{table}[!t]
  \centering
  \small
  \caption{\textbf{Ablation of Multi-Scale Spatial Aggregation.} Different combinations of input resolutions are evaluated to quantify the contribution of cross-scale feature aggregation.}
  \label{tab:3}
  \renewcommand{\arraystretch}{0.75}
  \begin{tabular}{ccc}
    \toprule
    Scales & mAcc & mAP \\
    \midrule
    $224^2$ & 90.43 & 99.78 \\
    $224^2$ \& $112^2$ & 90.86 ($\uparrow$0.43) & 99.32 ($\downarrow$0.46) \\
    $224^2$ \& $448^2$ & 93.28 ($\uparrow$2.85) & 99.07 ($\downarrow$0.71) \\
    \rowcolor{blue!20}
    $224^2$ \& $112^2$ \& $448^2$ & \textbf{96.01} ($\uparrow$5.58) & \textbf{99.96} ($\uparrow$0.18) \\
    \bottomrule
  \end{tabular}
\end{table}

\begin{table}[!t]
  \centering
  \small
  \caption{\textbf{Ablation of Wavelet-Enhanced Central Stem.} Wavelet and Central denote the wavelet-enhancement Central Stem and central correlation convolution, respectively.}
  \label{tab:4}
  \renewcommand{\arraystretch}{0.87}
  \begin{tabular}{cccc}
    \toprule
    Wavelet & Central & mAcc & mAP \\
    \midrule
    $\times$ & $\times$ & 88.37 & 95.24 \\
    $\times$ & $\checkmark$ & 92.03 ($\uparrow$3.66) & 99.91 ($\uparrow$4.67) \\
    $\checkmark$ & $\times$ & 89.46 ($\uparrow$1.09) & 97.93 ($\uparrow$2.69) \\
    \rowcolor{blue!20}
    $\checkmark$ & $\checkmark$ & \textbf{96.01} ($\uparrow$7.64) & \textbf{99.96} ($\uparrow$4.72) \\
    \bottomrule
  \end{tabular}
\end{table}

\begin{table*}[!ht]
\caption{\textbf{Robustness comparison under common image perturbations ($\%$).} Each row reports accuracy under one perturbation setting or their combination. Average Drop denotes the mean accuracy decrease relative to the clean test benchmark.}
 \centering
 \tabcolsep=0.3cm
 \resizebox{\textwidth}{!}{
    \begin{tabular}{cccc>{\columncolor{blue!20}}ccccccc}
    \bottomrule \hline
    w/Blur & w/Cropping & w/JPEG & w/Noise & Ours & FreqNet & FerretNet & NPR & SAFE & ResNet-50 & UFD\\
\bottomrule \hline
$\checkmark$&$\times$&$\times$&$\times$&\textbf{95.96}&75.93&64.63&62.55&49.02&59.84&68.69\\
$\times$&$\checkmark$&$\times$&$\times$&\textbf{94.74}&84.29&73.48&69.72&50.12&60.77&70.19\\
$\times$&$\times$&$\checkmark$&$\times$&\textbf{95.99}&86.24&72.13&70.89&50.80&61.26&69.61\\
$\times$&$\times$&$\times$&$\checkmark$&\textbf{93.39}&84.94&67.41&66.51&50.08&60.30&68.45\\
$\checkmark$&$\checkmark$&$\checkmark$&$\checkmark$&\textbf{95.01}&82.95&69.29&67.10&50.40&60.42&69.11\\
\bottomrule
\multicolumn{4}{c}{\textbf{Average Drop}} 
& \textbf{$\downarrow$ 0.99} & $\downarrow$ 3.64 & $\downarrow$ 9.48 & $\downarrow$ 5.87 & $\downarrow$ 14.03 & $\downarrow$ 1.79 &$\downarrow$ 1.21\\
\bottomrule
\end{tabular}
}
\label{tab:5}
\end{table*}

\begin{table}[!t]
  \centering
  \small
  \caption{\textbf{Ablation of Frequency-Aware Prediction Block.} FFConv and Gate denote the Fast Fourier Convolution and gated operation, respectively.}
  \label{tab:freq_ablation}
  \renewcommand{\arraystretch}{0.87}
  \begin{tabular}{cccc}
    \toprule
    FFConv & Gate & mAcc & mAP \\
    \midrule
    $\times$ & $\times$ & 81.98 & 96.97 \\
    $\checkmark$ & $\times$ & 85.06 ($\uparrow$3.08) & 99.17 ($\uparrow$2.20) \\
    $\times$ & $\checkmark$ & 86.02 ($\uparrow$4.04) & 99.58 ($\uparrow$2.61) \\
    \rowcolor{blue!20}
    $\checkmark$ & $\checkmark$ & \textbf{96.01} ($\uparrow$14.03) & \textbf{99.96} ($\uparrow$2.99) \\
    \bottomrule
  \end{tabular}
\end{table}

\subsection{Main Results}
Tabs. \ref{tab:1} and \ref{tab:2} report accuracy and average precision on the CTForensics test benchmark. ESF-CTFD achieves the best overall performance, with 96.01\% mean accuracy and 99.96\% mean average precision. Compared with the strongest existing methods, FreqNet, ESF-CTFD improves mean accuracy by 9.50\% and mean average precision by 1.07\%. The improvement indicates that ESF-CTFD benefits from jointly modeling local wavelet-enhanced residual cues and global frequency artifacts, which are complementary for detecting diverse AI-generated CT images.

The per-generator results show that ESF-CTFD generalizes beyond the training generator. Although trained with HA-GAN as the generated source, it achieves near-perfect or perfect accuracy on CGAN, sRD-GAN, GenerateCT, LungDDPM, LungDDPM+, HA-GAN, and Text2CT. On CTGAN, MedSyn, and StableDiffusion, the accuracy is lower than on other sources, but the average precision remains high, reaching 99.81\%, 99.91\%, and 99.89\%, respectively. This suggests that ESF-CTFD learns robust discriminative cues for ranking authentic and generated CT slices, even when unseen generators produce samples with more ambiguous decision margins.

The comparison also reveals a clear domain gap for detectors adapted from natural-image forensics. SAFE and ResNet-50 obtain strong results on specific generators such as MedSyn or HA-GAN, but their mean accuracy remains limited. Frequency-oriented methods \cite{li2021frequency} are more competitive: FreqNet obtains 86.51\% mean accuracy and 98.89\% mean average precision, confirming that spectral artifacts are informative for CT forensics. However, ESF-CTFD consistently benefits from combining wavelet-enhanced local cues, multi-scale anatomical aggregation, and frequency-aware prediction, leading to more balanced performance across heterogeneous GAN- and diffusion-based sources.

\subsection{Ablation Study}
Tab. \ref{tab:3} evaluates the effect of Multi-Scale Spatial Aggregation. Using only the $224 \times 224$ branch gives 90.43\% mean accuracy. Adding the lower-resolution $112 \times 112$ branch provides a small accuracy gain, while adding the higher-resolution $448 \times 448$ branch improves accuracy to 93.28\%. The full three-scale setting reaches 96.01\% mean accuracy and 99.96\% mean average precision, showing that local high-resolution evidence and coarse structural context are complementary for generated CT image detection.

Tab. \ref{tab:4} studies the Wavelet-Enhanced Central Stem. The central correlation branch alone improves mean accuracy by 3.66\%, suggesting that local grayscale intensity relationships are strong cues for CT forgery detection. The wavelet branch alone brings a smaller but consistent improvement by enhancing directional high-frequency residuals. Combining both branches yields the best result, improving mean accuracy by 7.64\% and mean average precision by 4.72\% over the baseline stem. This supports the design choice of coupling wavelet-domain enhancement with local central-difference modeling.

Tab. \ref{tab:freq_ablation} ablates the Frequency-Aware Prediction Block. Fast Fourier Convolution improves mean accuracy by 3.08\% and mean average precision by 2.20\%, showing the benefit of global spectral modeling. The gate yields a larger gain, indicating that gated feature selection helps suppress weak responses. Combining Fast Fourier Convolution with the gate achieves the best performance, improving mean accuracy by 14.03\% and mean average precision by 2.99\%. This confirms that spectral refinement and gated selection are complementary for distinguishing generated CT slices.

\subsection{Robustness Analysis}
We further evaluate robustness under common image degradations. Following \cite{frank2020leveraging}, five perturbation settings are used, including Gaussian blur, random cropping, JPEG compression, Gaussian noise, and their combination. For each selected perturbation, every image is perturbed with a probability of 50\%. Specifically, Gaussian blur uses a kernel size randomly sampled from $\{3,5,7,9\}$; random cropping samples the cropping percentage from $U(5,20)$ and upsamples the cropped region back to the original size; JPEG compression samples the quality factor from $U(10,75)$; and Gaussian noise samples the variance from $U(5.0,20.0)$.

As shown in Tab. \ref{tab:5}, ESF-CTFD remains stable across all perturbation settings. The model obtains 95.96\% accuracy under blur, 94.74\% under cropping, 95.99\% under JPEG compression, 93.39\% under Gaussian noise, and 95.01\% when all perturbations are combined. The average drop is only 0.99\%, which is lower than the drops of FreqNet, FerretNet, NPR, and SAFE. ResNet-50 and UFD show small average drops, but their absolute accuracies remain much lower, indicating that low degradation sensitivity alone does not imply reliable detection. The strong robustness of ESF-CTFD comes from preserving complementary local, cross-scale, and spectral forgery cues under input distortions.

\section{Conclusion}

In this paper, we present CTForensics, a comprehensive dataset for AI-generated CT image detection that supports generalization evaluation across ten representative generative models and 75,990 images. We further propose ESF-CTFD, a CT-oriented detector that integrates a Wavelet-Enhanced Central Stem, Multi-Scale Spatial Aggregation, and a Frequency-Aware Prediction Block to model local residual, anatomical, and spectral forgery cues. Extensive experiments show that ESF-CTFD outperforms existing detectors and maintains strong generalization under both clean and perturbed test settings. These findings highlight the importance of CT-specific forensic modeling for trustworthy medical image analysis.
\textbf{Future Work.} In the future, we will extend CTForensics to 3D volume-level detection to model inter-slice consistency and anatomical coherence. We will also include more CT generative models and clinical deployment scenarios to further evaluate real-world robustness.

\bibliographystyle{IEEEtran}
\bibliography{refs}

@inproceedings{ojha2023towards,
  title={Towards universal fake image detectors that generalize across generative models},
  author={Ojha, Utkarsh and Li, Yuheng and Lee, Yong Jae},
  booktitle={Proceedings of the IEEE/CVF Conference on Computer Vision and Pattern Recognition},
  pages={24480--24489},
  year={2023}
}

@article{loey2025deep,
  title={A deep transfer learning model with classical data augmentation and CGAN to detect COVID-19 from chest CT radiography digital images},
  author={Loey, Mohamed and Manogaran, Gunasekaran and Khalifa, Nour Eldeen M},
  journal={Neural computing and applications},
  volume={37},
  number={35},
  pages={29099--29111},
  year={2025},
  publisher={Springer}
}

@inproceedings{mirsky2019ct,
  title={$\{$CT-GAN$\}$: Malicious tampering of 3d medical imagery using deep learning},
  author={Mirsky, Yisroel and Mahler, Tom and Shelef, Ilan and Elovici, Yuval},
  booktitle={28th USENIX Security Symposium (USENIX Security 19)},
  pages={461--478},
  year={2019}
}

@article{lee2022diverse,
  title={Diverse covid-19 CT image-to-image translation with stacked residual dropout},
  author={Lee, Kin Wai and Chin, Renee Ka Yin},
  journal={Bioengineering},
  volume={9},
  number={11},
  pages={698},
  year={2022},
  publisher={MDPI}
}

@inproceedings{hamamci2024generatect,
  title={GenerateCT: Text-conditional generation of 3D chest CT volumes},
  author={Hamamci, Ibrahim Ethem and Er, Sezgin and Sekuboyina, Anjany and Simsar, Enis and Tezcan, Alperen and Simsek, Ayse Gulnihan and Esirgun, Sevval Nil and Almas, Furkan and Do{\u{g}}an, Irem and Dasdelen, Muhammed Furkan and others},
  booktitle={European Conference on Computer Vision},
  pages={126--143},
  year={2024},
  organization={Springer}
}

@article{jiang2025lung,
  title={Lung-ddpm: Semantic layout-guided diffusion models for thoracic ct image synthesis},
  author={Jiang, Yifan and Lemar{\'e}chal, Yannick and Bafaro, Jos{\'e}e and Abi-Rjeile, Jessica and Joubert, Philippe and Despr{\'e}s, Philippe and Manem, Venkata},
  journal={arXiv preprint arXiv:2502.15204},
  year={2025}
}

@article{jiang2025lung+,
  title={Lung-DDPM+: Efficient thoracic CT image synthesis using diffusion probabilistic model},
  author={Jiang, Yifan and Shariftabrizi, Ahmad and Manem, Venkata SK},
  journal={Computers in biology and medicine},
  volume={199},
  pages={111290},
  year={2025},
  publisher={Elsevier}
}

@article{xu2024medsyn,
  title={MedSyn: text-guided anatomy-aware synthesis of high-fidelity 3-D CT images},
  author={Xu, Yanwu and Sun, Li and Peng, Wei and Jia, Shuyue and Morrison, Katelyn and Perer, Adam and Zandifar, Afrooz and Visweswaran, Shyam and Eslami, Motahhare and Batmanghelich, Kayhan},
  journal={IEEE Transactions on Medical Imaging},
  volume={43},
  number={10},
  pages={3648--3660},
  year={2024},
  publisher={IEEE}
}

@inproceedings{konz2024anatomically,
  title={Anatomically-controllable medical image generation with segmentation-guided diffusion models},
  author={Konz, Nicholas and Chen, Yuwen and Dong, Haoyu and Mazurowski, Maciej A},
  booktitle={International Conference on Medical Image Computing and Computer-Assisted Intervention},
  pages={88--98},
  year={2024},
  organization={Springer}
}

@article{grabovski2025back,
  title={Back-in-time diffusion: Unsupervised detection of medical deepfakes},
  author={Grabovski, Fred M and Yasur, Lior and Amit, Guy and Mirsky, Yisroel},
  journal={ACM Transactions on Intelligent Systems and Technology},
  volume={16},
  number={6},
  pages={1--26},
  year={2025},
  publisher={ACM New York, NY}
}

@article{molino2025text,
  title={Text-to-CT Generation via 3D Latent Diffusion Model with Contrastive Vision-Language Pretraining},
  author={Molino, Daniele and Caruso, Camillo Maria and Ruffini, Filippo and Soda, Paolo and Guarrasi, Valerio},
  journal={arXiv preprint arXiv:2506.00633},
  year={2025}
}

@article{hamamci2024developing,
  title={Developing generalist foundation models from a multimodal dataset for 3d computed tomography},
  author={Hamamci, Ibrahim Ethem and Er, Sezgin and Wang, Chenyu and Almas, Furkan and Simsek, Ayse Gulnihan and Esirgun, Sevval Nil and Dogan, Irem and Durugol, Omer Faruk and Hou, Benjamin and Shit, Suprosanna and others},
  journal={arXiv preprint arXiv:2403.17834},
  year={2024}
}

@inproceedings{rombach2022high,
  title={High-resolution image synthesis with latent diffusion models},
  author={Rombach, Robin and Blattmann, Andreas and Lorenz, Dominik and Esser, Patrick and Ommer, Bj{\"o}rn},
  booktitle={Proceedings of the IEEE/CVF conference on computer vision and pattern recognition},
  pages={10684--10695},
  year={2022}
}

@article{goodfellow2014generative,
  title={Generative adversarial nets},
  author={Goodfellow, Ian J and Pouget-Abadie, Jean and Mirza, Mehdi and Xu, Bing and Warde-Farley, David and Ozair, Sherjil and Courville, Aaron and Bengio, Yoshua},
  journal={Advances in neural information processing systems},
  volume={27},
  year={2014}
}

@article{song2020score,
  title={Score-based generative modeling through stochastic differential equations},
  author={Song, Yang and Sohl-Dickstein, Jascha and Kingma, Diederik P and Kumar, Abhishek and Ermon, Stefano and Poole, Ben},
  journal={arXiv preprint arXiv:2011.13456},
  year={2020}
}

@article{touati2021feature,
  title={A feature invariant generative adversarial network for head and neck MRI/CT image synthesis},
  author={Touati, Redha and Le, William Trung and Kadoury, Samuel},
  journal={Physics in Medicine \& Biology},
  volume={66},
  number={9},
  pages={095001},
  year={2021},
  publisher={IOP Publishing}
}

@article{lyu2022conversion,
  title={Conversion between CT and MRI images using diffusion and score-matching models},
  author={Lyu, Qing and Wang, Ge},
  journal={arXiv preprint arXiv:2209.12104},
  year={2022}
}

@article{zhu2024generative,
  title={Generative enhancement for 3d medical images},
  author={Zhu, Lingting and Codella, Noel and Chen, Dongdong and Jin, Zhenchao and Yuan, Lu and Yu, Lequan},
  journal={arXiv preprint arXiv:2403.12852},
  year={2024}
}

@inproceedings{shao2025trace,
  title={TRACE: Temporally Reliable Anatomically-Conditioned 3D CT Generation with Enhanced Efficiency},
  author={Shao, Minye and Miao, Xingyu and Duan, Haoran and Wang, Zeyu and Chen, Jingkun and Huang, Yawen and Wu, Xian and Deng, Jingjing and Long, Yang and Zheng, Yefeng},
  booktitle={International Conference on Medical Image Computing and Computer-Assisted Intervention},
  pages={627--637},
  year={2025},
  organization={Springer}
}

@inproceedings{li2025toward,
  title={Toward Medical Deepfake Detection: A Comprehensive Dataset and Novel Method},
  author={Li, Shuaibo and Xing, Zhaohu and Wang, Hongqiu and Hao, Pengfei and Li, Xingyu and Liu, Zekai and Zhu, Lei},
  booktitle={International Conference on Medical Image Computing and Computer-Assisted Intervention},
  pages={626--637},
  year={2025},
  organization={Springer}
}

@inproceedings{he2016deep,
  title={Deep residual learning for image recognition},
  author={He, Kaiming and Zhang, Xiangyu and Ren, Shaoqing and Sun, Jian},
  booktitle={Proceedings of the IEEE conference on computer vision and pattern recognition},
  pages={770--778},
  year={2016}
}

@article{albahli2024mednet,
  title={MedNet: Medical deepfakes detection using an improved deep learning approach},
  author={Albahli, Saleh and Nawaz, Marriam},
  journal={Multimedia Tools and Applications},
  volume={83},
  number={16},
  pages={48357--48375},
  year={2024},
  publisher={Springer}
}

@inproceedings{tan2021efficientnetv2,
  title={Efficientnetv2: Smaller models and faster training},
  author={Tan, Mingxing and Le, Quoc},
  booktitle={International conference on machine learning},
  pages={10096--10106},
  year={2021},
  organization={PMLR}
}

@article{solaiyappan2022machine,
  title={Machine learning based medical image deepfake detection: A comparative study},
  author={Solaiyappan, Siddharth and Wen, Yuxin},
  journal={Machine Learning with Applications},
  volume={8},
  pages={100298},
  year={2022},
  publisher={Elsevier}
}

@article{prezja2022deepfake,
  title={DeepFake knee osteoarthritis X-rays from generative adversarial neural networks deceive medical experts and offer augmentation potential to automatic classification},
  author={Prezja, Fabi and Paloneva, Juha and P{\"o}l{\"o}nen, Ilkka and Niinim{\"a}ki, Esko and {\"A}yr{\"a}m{\"o}, Sami},
  journal={Scientific Reports},
  volume={12},
  number={1},
  pages={18573},
  year={2022},
  publisher={Nature Publishing Group UK London}
}

@article{sun2022hierarchical,
  title={Hierarchical amortized GAN for 3D high resolution medical image synthesis},
  author={Sun, Li and Chen, Junxiang and Xu, Yanwu and Gong, Mingming and Yu, Ke and Batmanghelich, Kayhan},
  journal={IEEE journal of biomedical and health informatics},
  volume={26},
  number={8},
  pages={3966--3975},
  year={2022},
  publisher={IEEE}
}

@inproceedings{tan2024rethinking,
  title={Rethinking the up-sampling operations in cnn-based generative network for generalizable deepfake detection},
  author={Tan, Chuangchuang and Zhao, Yao and Wei, Shikui and Gu, Guanghua and Liu, Ping and Wei, Yunchao},
  booktitle={Proceedings of the IEEE/CVF Conference on Computer Vision and Pattern Recognition},
  pages={28130--28139},
  year={2024}
}

@inproceedings{li2025improving,
  title={Improving synthetic image detection towards generalization: An image transformation perspective},
  author={Li, Ouxiang and Cai, Jiayin and Hao, Yanbin and Jiang, Xiaolong and Hu, Yao and Feng, Fuli},
  booktitle={Proceedings of the 31st ACM SIGKDD Conference on Knowledge Discovery and Data Mining V. 1},
  pages={2405--2414},
  year={2025}
}

@inproceedings{chollet2017xception,
  title={Xception: Deep learning with depthwise separable convolutions},
  author={Chollet, Fran{\c{c}}ois},
  booktitle={Proceedings of the IEEE conference on computer vision and pattern recognition},
  pages={1251--1258},
  year={2017}
}

@inproceedings{finder2024wavelet,
  title={Wavelet convolutions for large receptive fields},
  author={Finder, Shahaf E and Amoyal, Roy and Treister, Eran and Freifeld, Oren},
  booktitle={European Conference on Computer Vision},
  pages={363--380},
  year={2024},
  organization={Springer}
}

@inproceedings{zhang2025cwnet,
  title={Cwnet: Causal wavelet network for low-light image enhancement},
  author={Zhang, Tongshun and Liu, Pingping and Lu, Yubing and Cai, Mengen and Zhang, Zijian and Zhang, Zhe and Zhou, Qiuzhan},
  booktitle={Proceedings of the IEEE/CVF International Conference on Computer Vision},
  pages={8789--8799},
  year={2025}
}

@inproceedings{yu2020searching,
  title={Searching central difference convolutional networks for face anti-spoofing},
  author={Yu, Zitong and Zhao, Chenxu and Wang, Zezheng and Qin, Yunxiao and Su, Zhuo and Li, Xiaobai and Zhou, Feng and Zhao, Guoying},
  booktitle={Proceedings of the IEEE/CVF conference on computer vision and pattern recognition},
  pages={5295--5305},
  year={2020}
}

@inproceedings{Chi2020FastFC,
  title={Fast Fourier Convolution},
  author={Lu Chi and Borui Jiang and Yadong Mu},
  booktitle={Neural Information Processing Systems},
  year={2020},
}

@inproceedings{li2021frequency,
  title={Frequency-aware discriminative feature learning supervised by single-center loss for face forgery detection},
  author={Li, Jiaming and Xie, Hongtao and Li, Jiahong and Wang, Zhongyuan and Zhang, Yongdong},
  booktitle={Proceedings of the IEEE/CVF conference on computer vision and pattern recognition},
  pages={6458--6467},
  year={2021}
}

@inproceedings{tan2024frequency,
  title={Frequency-aware deepfake detection: Improving generalizability through frequency space domain learning},
  author={Tan, Chuangchuang and Zhao, Yao and Wei, Shikui and Gu, Guanghua and Liu, Ping and Wei, Yunchao},
  booktitle={Proceedings of the AAAI Conference on Artificial Intelligence},
  volume={38},
  number={5},
  pages={5052--5060},
  year={2024}
}

@inproceedings{zheng2024breaking,
 author = {Zheng, Chende and Lin, Chenhao and Zhao, Zhengyu and Wang, Hang and Guo, Xu and Liu, Shuai and Shen, Chao},
 booktitle = {Advances in Neural Information Processing Systems},
 pages = {59570--59596},
 title = {Breaking Semantic Artifacts for Generalized AI-generated Image Detection},
 volume = {37},
 year = {2024}
}

@inproceedings{
liang2025ferretnet,
title={FerretNet: Efficient Synthetic Image Detection via Local Pixel Dependencies},
author={Shuqiao Liang and Jian Liu and Chen Renzhang and Quanlong Guan},
booktitle={The Thirty-ninth Annual Conference on Neural Information Processing Systems},
year={2025},
}

@inproceedings{frank2020leveraging,
  title={Leveraging frequency analysis for deep fake image recognition},
  author={Frank, Joel and Eisenhofer, Thorsten and Sch{\"o}nherr, Lea and Fischer, Asja and Kolossa, Dorothea and Holz, Thorsten},
  booktitle={International conference on machine learning},
  pages={3247--3258},
  year={2020},
  organization={PMLR}
}

@article{van2008visualizing,
  title={Visualizing data using t-SNE.},
  author={Van der Maaten, Laurens and Hinton, Geoffrey},
  journal={Journal of machine learning research},
  volume={9},
  number={11},
  year={2008}
}

@inproceedings{li2026towards,
  title={Towards generalizable ai-generated image detection via image-adaptive prompt learning},
  author={Li, Yiheng and Tan, Zichang and Xu, Guoqing and Lei, Zhen and Zhou, Xu and Yang, Yang},
  booktitle={Proceedings of the IEEE/CVF Conference on Computer Vision and Pattern Recognition},
  pages={21262--21272},
  year={2026}
}

\end{document}